\pdfoutput=1

\documentclass[11pt]{article}

\usepackage{CJKutf8}
\usepackage{graphicx}

\usepackage{booktabs} 
\usepackage{soul}
\usepackage{amsmath}
\usepackage{color, xcolor}
\soulregister\cite7 
\soulregister\citep7 
\soulregister\citet7 
\soulregister\ref7 
\soulregister\pageref7 
\usepackage{multirow}
\usepackage{booktabs}
\usepackage{emnlp2021}
\usepackage{graphicx} 
\usepackage{float} 
\usepackage{subfigure} 
\usepackage{amsmath,amssymb,amsfonts}
\usepackage{times}
\usepackage{latexsym}
\usepackage{hyperref}
\usepackage[T1]{fontenc}

\usepackage[utf8]{inputenc}

\usepackage{microtype}

\title{An Error-Guided Correction Model for Chinese Spelling Error Correction}
\author{Rui Sun$^{1,2}$,
        Xiuyu Wu$^{1,2}$, 
        Yunfang Wu$^{1,3}$\thanks{~~Corresponding author.} \\
    $^{1}$MOE Key Laboratory of Computational Linguistics, Peking University Beijing, China\\ 
    $^{2}$School of Software and Microelectronics, Peking University, Beijing, China \\
    $^{3}$School of Computer Science, Peking University, Beijing, China \\
    \texttt{\{sunrui0720, xiuyu\_wu\}@stu.pku.edu.cn},
    \texttt{wuyf@pku.edu.cn}
    }

\begin{document}
\begin{CJK}{UTF8}{gbsn}
\maketitle


\begin{abstract}
Although existing neural network approaches have achieved great success on Chinese spelling correction, there is still room to improve. The model is required to avoid over-correction and to distinguish a correct token from its phonological and visually similar ones. In this paper, we propose an error-guided correction model (EGCM) to improve Chinese spelling correction. By borrowing the powerful ability of BERT, we propose a novel zero-shot error detection method to do a preliminary detection, which guides our model to attend more on the probably wrong tokens in encoding and to avoid modifying the correct tokens in generating.
Furthermore, we introduce a new loss function to integrate the error confusion set, which enables our model  
to distinguish easily misused tokens. Moreover, our model supports highly parallel decoding to meet real application requirements. 
Experiments are conducted on widely used benchmarks. 
Our model achieves superior performance against state-of-the-art approaches by a remarkable margin, on both the correction quality and computation speed.  
Our codes are publicly available at \href{https://github.com/ruisun1/Mask-Predict-main}{https://github.com/ruisun1/Mask-Predict-main}
\end{abstract}

\section{Introduction}

Chinese spelling correction (CSC) attracts wide attention in recent years, which is significant for many  
real applications, such as search engine \cite{conf/tal/MartinsS04}, optical character recognition(OCR) \cite{conf/lrec/AfliQWS16} and automatic speech recognition(ASR) \cite{hinton2012deep}. 

\begin{figure}[t] 
    \includegraphics[width=0.5\textwidth]{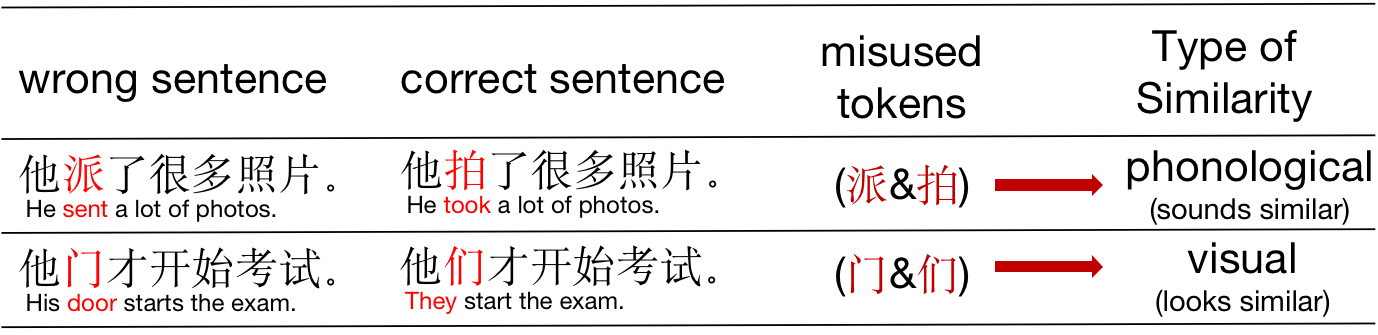} 
    \caption{Examples of Chinese spelling errors. Misspelled characters and their corresponding corrections are marked in red.} 
    \label{example} 
\end{figure}

Given an input sentence with spelling errors, the model is trained to detect and correct these errors and output a correct sentence. According to \citet{2010Visually}, phonologically and visually similar characters are major contributing factors for errors in Chinese text. As shown in Figure \ref{example}, in the first example, the error is caused by the misuse of "派"(send) and "拍"(take) which have similar Chinese pronunciation.
In the second example, the error is caused by the misuse of "门"(door) and "们"(they) which have similar shapes. 

Recently, the advanced neural network models and pre-trained models have achieved great success in CSC, such as PLOME \cite{liu-etal-2021-plome}, REALISE \cite{2021Read}, PHMOSpell \cite{2021PHMOSpell}, SpellBert \cite{ji2021spellbert} , GAD \cite{guo2021global}, MLM-phonetics \cite{journals/corr/abs-2004-14166}, RoBERTa-DCN \cite{wang2021dynamic} and ECSpell \cite{2022General}. Although much progress has been made, there are still limitations in previous methods. 

First, given an input sequence, only a small fragment might be misspelled. However, for most of the previous models, they are totally blind to the errors at start, and so they attend on all tokens equally in encoding and generate every token from left to right for inference. As a result, previous models are inefficient and might create over-correction. As these models obtain a stronger ability to correct the errors, 
they also tend to modify the correct tokens by mistake.



\begin{figure}[t] 
    \centering
    \includegraphics[width=7cm]{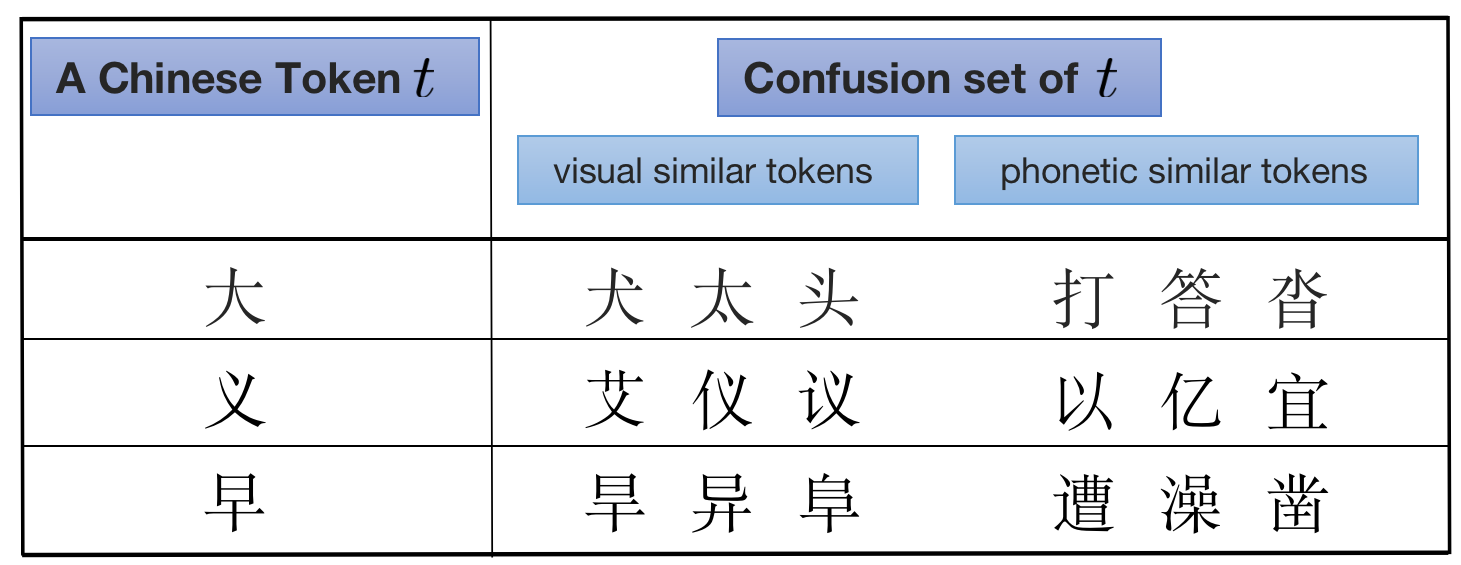} 
    \caption{An example of the confusion set.} 
    \label{confusion} 
\end{figure}

Second, the confusion set, where a set of phonological and visual similar tokens are defined for each Chinese token, provides valuable knowledge for spelling correction, as shown in Figure \ref{confusion}. But the methodology to use it should be further improved. 
For example, \citet{liu-etal-2021-plome} propose a Confusion Set based Masking Strategy, in which they remove a token and replace the token with a random character in the confusion set. 
As the model randomly chooses a token from the confusion set each time, some tokens might be ignored. Besides,
this method can't pay more attention to the token that is more easily to be misused. 
\citet{conf/acl/WangTZ19} propose to generate a character from the confusion set rather than the entire vocabulary. In this hard restriction, the model cannot generate tokens that are not in the confusion set. 

Third, when a CSC model is deployed in real applications, the time cost of inference is a critical problem to be considered. However, most previous models try to improve the generation quality but ignore the computation speed. 

To address these issues mentioned above, we propose an Error-Guided Correction Model (EGCM) for CSC. Firstly, taking advantage of the strong ability of BERT \cite{2018BERT}, we propose a novel zero-shot error detection method to do a preliminary detection, which  provides precise guidance signals to the correction model. Following the guidance, our model attends more on the probably wrong tokens in encoding, and fixes the probably correct tokens during generation to avoid over-correction. Furthermore, we introduce a new loss function that effectively integrates the confusion set. 
By applying this loss function, every similar token in the confusion set is learned to be distinguished from the target token, and the most similar token with a high possibility of being misused is given more attention.
To speed up the inference, we apply a mask-predict strategy \cite{conf/emnlp/GhazvininejadLL19} to support parallel decoding, where the tokens with low generation probability
are masked and predicted iteratively. 

We conduct extensive experiments on the widely used benchmark dataset SIGHAN \cite{conf/acl-sighan/WuLL13,conf/acl-sighan/YuL14,conf/acl-sighan/TsengLCC15}. Experimental results show that our model significantly outperforms all 
previous approaches, achieving a new state-of-the-art performance for Chinese spelling correction. Moreover, 
our model has a distinct speed advantage over other models, which is 6.3 times faster than the standard Transformer and 1.5 times faster than the recent non-autoregressive model TtT \cite{2021Tail}.

We summarize our contributions as follows:
\begin{itemize}
    \item We propose a novel zero-shot error detection method,
    which guides the correction model to attend more on the probably wrong tokens in encoding and fix the probably correct tokens in inference to avoid over-correction.
    \item We propose a new loss function to take advantage of the confusion set, which enables our model to distinguish similar tokens and attach more importance to the easily misused tokens.
    \item We apply an error-guided mask-predict decoding strategy for spelling correction, which supports highly parallel decoding and greatly accelerates the computation speed.
    \item We integrate all modules into a unified model, which achieves a new state-of-the-art performance for both correction quality and inference speed.
\end{itemize}

\begin{figure*}[t] 
    \centering 
    \includegraphics[width=1\textwidth]{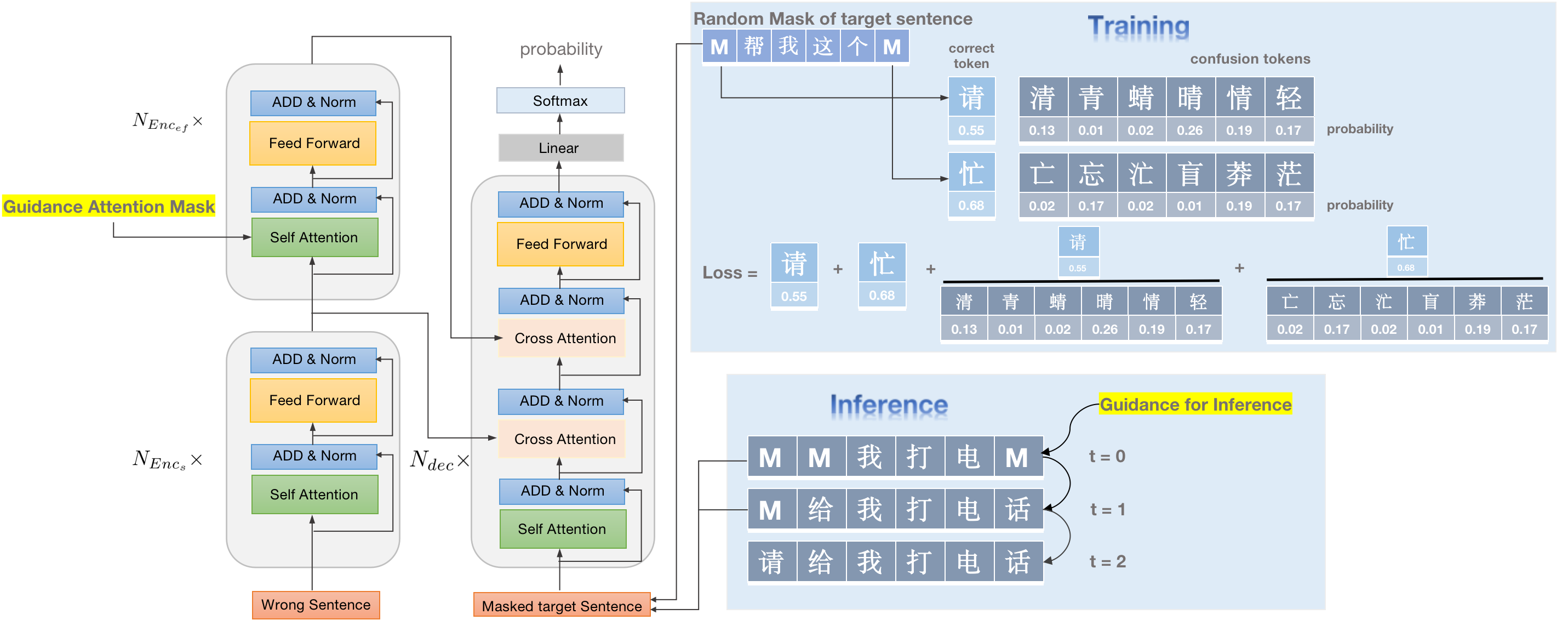} 
    \caption{The architecture of our proposed model. "M" denotes the \texttt{[MASK]} token. \textit{Guidance for Inference} and \textit{Guidance Attention Mask} are generated from Zero-shot error detection as shown in Figure \ref{zero-shot}. } 
    \label{Fig.main1} 
\end{figure*}

\section{Related work}
CSC is a task that detect and correct wrong tokens in Chinese Sentences. It's an active topic that varieties of approaches have been proposed to tackle the task \cite{conf/acl/WangTZ19,journals/corr/abs-2004-14166,2021Tail,2021Read,liu-etal-2021-plome,2021PHMOSpell}.  

\begin{figure}[t] 
    \centering 
    \includegraphics[width=0.5\textwidth]{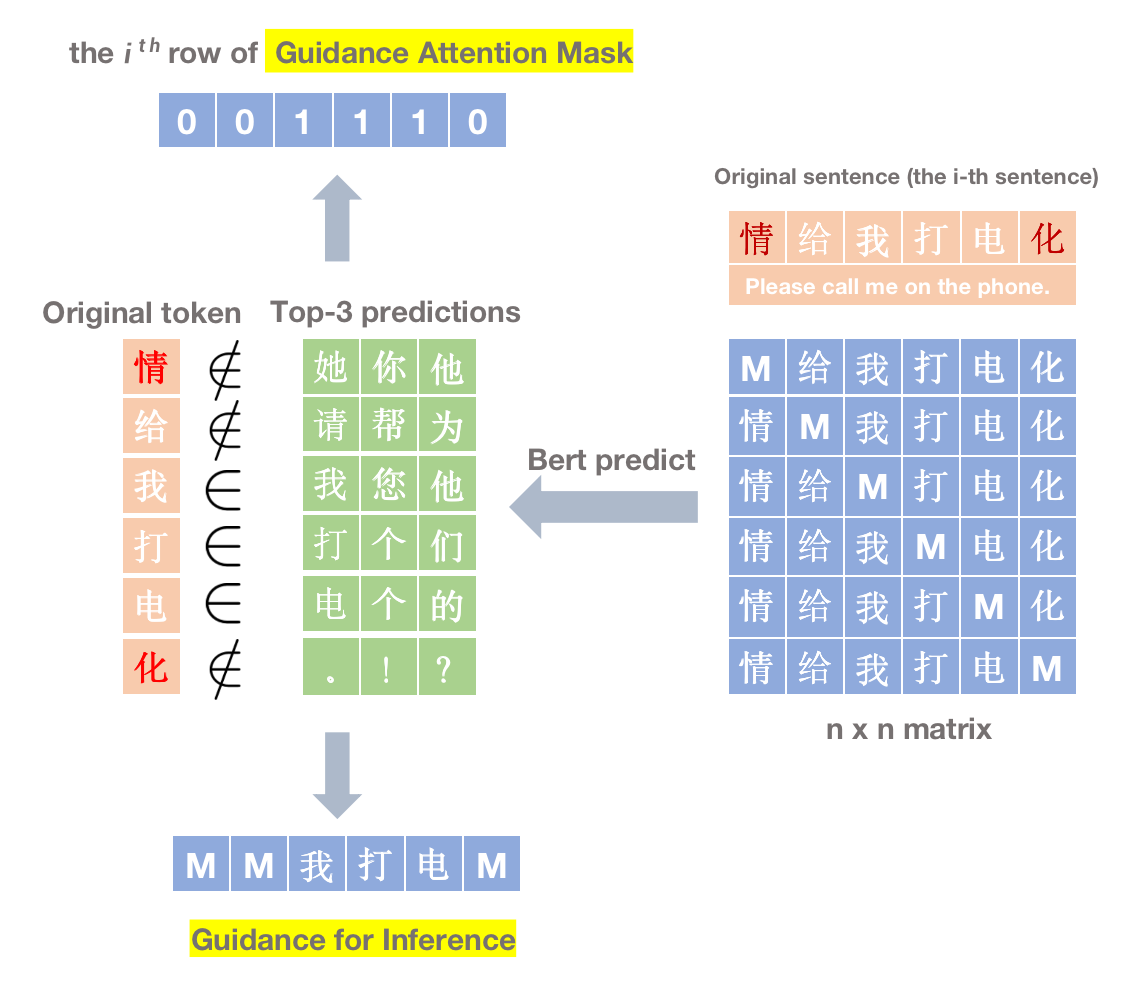} 
    \caption{An example of how our model obtains guidance signals using a zero-shot method. The original wrong token is marked red. } 
    \label{zero-shot} 
\end{figure}

Earlier work in CSC focuses mainly on unsupervised methods, which typically adopts a confusion set to find correct candidates and employs a language model to select the correct one \cite{conf/acl-sighan/ChenLLWC13,conf/acl-sighan/YuL14}. Recently, sequence translation and sequence tagging are the two most widely used methods in CSC. \citet{wang2018hybrid}  treats the CSC task as a sequence labeling problem, and use
a bidirectional LSTM to predict the correct characters. \citet{liu-etal-2021-plome, ji2021spellbert, 2021Read,2022General} try to enrich the representation generated by the encoder by introducing visual and phonetic features. Softmax operation is utilized to find a substitution for each token in the sentence. As the rapid development of neural machine translation \cite{vaswani2017attention}, seq2seq encoder-decoder frameworks have been introduced to the CSC task in \cite{journals/corr/JiWTGTG17,journals/corr/ChollampattTN16,conf/acl/WangTZ19}.

Recent work tends to utilize character similarity as an external knowledge. The confusion set where similar characters are stored is widely used \cite{liu-etal-2021-plome,conf/acl/ZhangHLL20,conf/acl/WangTZ19,conf/acl-sighan/YuL14,journals/corr/abs-2004-14166,2022General}. There are several ways of using the confusion set. The first is to augment the training data by replacing the original token with it's similar tokens \cite{liu-etal-2021-plome,conf/acl/ZhangHLL20}.  \citet{conf/acl/WangTZ19} proposes to generate a character from the confusion set rather than the entire vocabulary. \citet{conf/acl-sighan/YuL14} proposes to produce candidates by retrieving the confusion set and then filter them via language models. \citet{journals/corr/abs-2004-14166} uses similarity graphs derived from the confusion set and use graph convolution operation to absorb the information from neighboring characters in the graph.

\section{Methodology}
The proposed Error-Guided 
Correction Model (EGCM) is illustrated in Figure \ref{Fig.main1}. 
We apply the conditional masked language model (CMLM) \cite{conf/emnlp/GhazvininejadLL19} as a backbone, which is an encoder-decoder architecture trained with a masked language model
objective \cite{2018BERT,lample2019crosslingual}. In the CMLM architecture, the source wrong sentence with n tokens is denoted as $X=(x_{1},x_{2},x_{3},...,x_{n})$, the target sentence is denoted as $Y=(y_{1},y_{2},y_{3},...,y_{n})$. Several tokens in $Y$ are replaced with \texttt{[MASK]}. These masked tokens construct the set $Y_{mask}$. And the rest of the tokens in $Y$ that are unmasked construct the set of $Y_{obs}$. For Chinese spelling correction, given
a source sentence $X$ and the set of unmasked target tokens $Y_{obs}$, the objec is to predict the probability $P(y|X,Y_{obs})$ and generate token y for each $y \in Y_{mask}$.  

We first propose a zero-shot spelling error detection method to provide two guidance signals to the correction model, as shown in Figure \ref{zero-shot}. The first guidance signal is the \textit{Guidance Attention Mask} that is used in the error-focused encoder, in which the probably correct tokens are masked to push our model to attend more on the wrong tokens. The second guidance signal is the \textit{Guidance for Inference} that 
serves as the start of decoding to avoid modifying correct tokens by mistake. Moreover, we introduce a new loss function to take advantage of the confusion set. During inference, we apply an error-guided mask-predict strategy in which the correct tokens are fixed and the probably wrong tokens are masked and repredicted iteratively. 

\subsection{Zero-shot Error Detection}\label{zero}
Given a sentence $X=(x_{1},x_{2},x_{3},...,x_{n})$ that contains $n$ tokens, we 
want to make a preliminary decision on which tokens are probably wrong and which are correct. 

As shown in Figure \ref{zero-shot}, firstly, we construct a $n\times n$ matrix by repeating the original sentence $n$ times, 
where the $k^{th}$ token is masked in the $k^{th}$ row in the matrix ($k$ is from 1 to $n$). Then, we employ BERT \cite{2018BERT} to predict each masked position condition on the unmasked tokens in the same row.
Thus, for each position from $x_{1}$ to $x_{n}$ in the sentence $X$, 
we obtain the predicted tokens along with their probabilities. 
The tokens with the top-$k$ probabilities are selected as candidates of modification. We assume that if the original token $x_{i}$ occurs in the candidates list, the token is considered correct. Otherwise, the token is probably wrong and needs to be corrected. 

Based on the output of error detection, we construct two guidance signals namely \textit{Guidance Attention Mask} and \textit{Guidance for Inference}, as shown in Figure \ref{zero-shot}. The \textit{Guidance Attention Mask} (GAM) is a matrix 
constructed by:

\begin{small}
\begin{equation}
\label{gam}
    GAM_{ij} = 
    \begin{cases}
        0,  &\texttt{$x_{ij}$ is probably wrong} \\ 
        1,  &\texttt{$ \space \space \space \space \space \space \space \space \space \space \space \space \space \space \space \space \space \space otherwise $}
    \end{cases}
\end{equation}
\end{small}
where $x_{ij}$ denotes the $j^{th}$ token in the $i^{th}$ sentence. $GAM_{ij}$ denotes the element of the $i^{th}$ row and the $j^{th}$ colomn in GAM.
The \textit{Guidance for Inference} (GFI) is constructed by masking all the probably wrong tokens in the original sentence.  
Further, GAM will be projected into the error-focused encoder, and GFI will be utilized to initialize the decoder.

\subsection{Error-aware Encoder}
\label{ewe}

We adopt the Transformer \cite{vaswani2017attention} encoder-decoder framework for Chinese spelling correction. 
We deviate from the standard Transformer encoder by fusing an error-focused encoder, as shown in the left part of  
Figure \ref{Fig.main1}. $Encoder_{s}$ is a standard Transformer encoder, and on top of that we introduce an error-focused encoder $Encoder_{ef}$, which utilizes \textit{Guidance Attention Mask} to expose the probably wrong tokens and divert the attention of our model from the correct tokens. 
The output of $Encoder_{s}$ is input into the error-focused encoder $Encoder_{ef}$. The
\textit{Guidance Attention Mask}
is used as an extra attention mask
in calculating self-attention in $Encoder_{ef}$, which informs the model which error part of the sentence should be focused on.
Concretely, the output of the $Encoder_s$ and $Encoder_{ef}$ is calculated respectively as:

\begin{small}
\begin{equation}
\label{encx}
    H^{s} = {\rm Encoder_s}({\rm Embedding}(X))
\end{equation}
\begin{equation}
\label{encz}
    H^{ef} = {\rm Encoder_{ef}}(H^{s}, atten\_mask = GAM)
\end{equation}
\end{small}


\subsection{Integrating Error Confusion Set for Training}
\label{efs}
During training, the tokens in $Y_{mask}$ are randomly selected among the target correct sentence as shown in Figure \ref{Fig.main1}. To better fit the requirements of correcting both single-character errors and multi-character errors in Chinese spelling correction, we adopt two masking strategies, namely mask-separate and mask-range. In mask-separate, we first sample the number of masked tokens from a uniform distribution between [1, len($X$)],
and then randomly choose that number of tokens. For mask-range, we select $l \in [2,3]$, and randomly select a span with length $l$. 
We replace 
the tokens in $Y_{mask}$ with a special \texttt{[MASK]} token, which is the generation object of the model.

There are three attention blocks in the Transformer decoder layer. After the self-attention block, the decoder will first attend to $H^{s}$, the representation of the source wrong sentence. Then, the decoder will attend to $H^{ef}$, the representation of the sentence with correct tokens being masked. The output of the previous decoder layer is then input into the next decoder layer. 

\begin{small}
\begin{equation}
    H^{l}_{d_{1}} =  {\rm selfAttention}(H^{l-1})
\end{equation}
\begin{equation}
   H^{l}_{d_{2}} = {\rm Attention}( Q = H^{l}_{d_{1}},K = H^{s}, V= H^{s})
\end{equation}

\begin{equation}
   H^{l} = {\rm Attention}( Q = H^{l}_{d_{2}},K = H^{ef}, V= H^{ef})
\end{equation}
\end{small}
where $H^{0}= Embedding(Y_{obs})$. Q, K, V represents the Query, Key, Value matrix. $Y_{obs}$ is the set of unmasked tokens in the target sentence. The output probability distribution $P$ is generated from the decoder over the vocabulary $V$:

\begin{small}
\begin{equation}
    P = {\rm softmax}(H^{l}W+b)
\end{equation}
\end{small}
where $H^{l}\in\mathbb{R}^{t \times d}$,  $W\in\mathbb{R}^{d \times \lvert v \rvert}$, $b\in\mathbb{R}^{t \times \lvert v \rvert}$. $t$ denotes the sequence length. 

We optimize the model 
over every token in $Y_{mask}$. Besides the traditional loss function,  
we introduce a new loss to integrate 
the confusion set knowledge.

We employ Maximum Likelihood Estimation (MLE) to conduct parameter learning and utilize negative log-likelihood (NLL) as the loss function, 
which is computed as:

\begin{small}
\begin{equation}
    L_{nll} = -\sum_{y_{i}\in{Y_{mask}}} \log{P(y_{i}|X,Y_{obs})} 
\end{equation}
\end{small}
To make full use of the confusion set knowledge, we introduce a new loss function $L_{cs}$. We adopt the confusion set constructed by \citet{2022General}. For each token $y_{i}$ in $Y_{mask}$, we find out the set of the similar tokens of $y_{i}$ based on the confusion set, namely $Y_{conf}$. The tokens in $Y_{conf}$ are regarded as negative samples of $y_{i}$. We use these negative samples to help our model better learn the difference between the target token and its similar ones. The optimization objective for the confusion loss  $L_{cs}$ is defined as:

\begin{small}
\begin{equation}
    L_{cs} = - \sum_{y_{i}\in{Y_{mask}}} \frac{ \log{P(y_{i}|X,Y_{obs})} }{\sum_{y_{c}\in{Y_{conf}}}\log{P(y_{c}|X,Y_{obs})}}
    \label{eq2}
\end{equation}
\end{small}
where $y_{c}$ denotes the similar token of $y_{i}$ in the confusion set.

Overall, the final optimization objective of our model is:

\begin{small}
\begin{equation}
    L_{f} = L_{nll} + \gamma \times L_{cs}
\end{equation}
\end{small}
where $\gamma$ is a hyperparameter to balance two loss functions.

\subsection{Error-Guided Generation}
\label{3.4}

In the inference stage, we apply a mask-predict approach \cite{conf/emnlp/GhazvininejadLL19}, where the tokens with low probability are masked and predicted within a constant number of iterations.

\begin{figure}[t] 
    \includegraphics[width=0.5\textwidth]{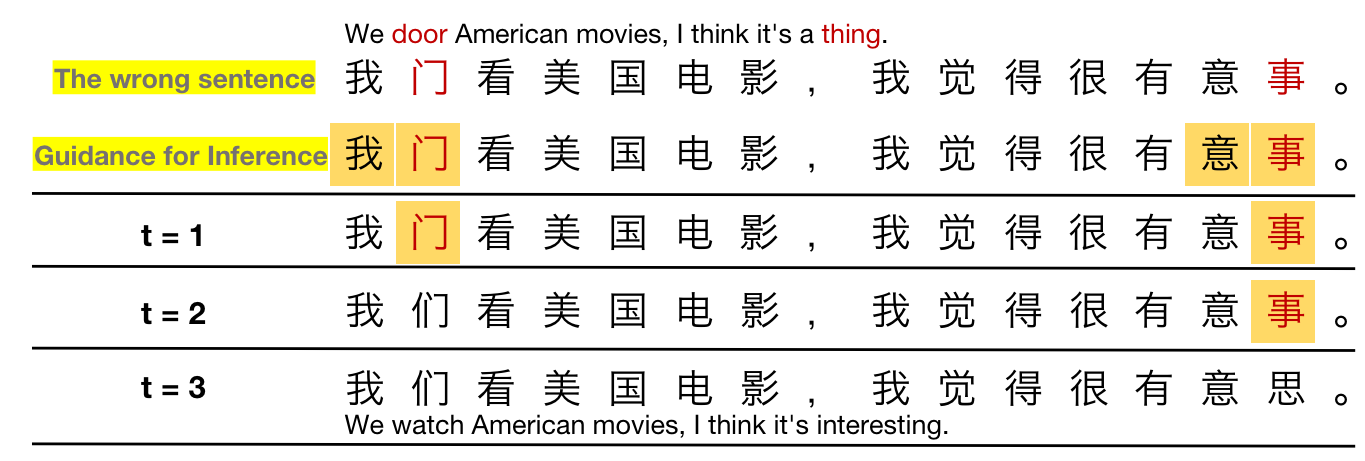} 
    \caption{An example of Error-Guided Generation. In Guidance for Inference, the masked tokens are highlighted. In later iterations, the highlighted tokens are of lowest probabilities and are masked and repredicted. The wrong tokens are marked in red. } 
    \label{Fig.main 3} 
\end{figure}

To provide the model a good start point for generation, we exploit
the \textit{Guidance for Inference} (GFI) 
 as an initialization for decoding. 
GFI  
produces a draft sentence, where the probably wrong tokens are masked and the probably correct ones are remained unmasked. During generation, the unmasked tokens will be fixed, and only the masked tokens are taken into consideration for modification in each iteration. Fixing these correct tokens will effectively teach our model to avoid over-correction.
Figure \ref{Fig.main 3} shows how does our model correct a wrong sentence in 3 iterations. 

The model runs for a pre-determined number of iterations $T$. The number of \texttt{[MASK]} in the 
draft sentence is denoted as $N_{ori}$. Accordingly,
the number of tokens that are masked in the $t_{th}$ iteration is defined as $N_{t}={N_{ori}}\times \frac{T-t}{T}$.
Formally, $Y_{mask}^{(0)}$ is the set of masked tokens in the \textit{Guidance for Inference}. At a later iteration $t$, we choose $N_{t}$ tokens among the masked tokens in the previous iteration $t-1$ that has the lowest probability scores:

\begin{small}
\begin{equation}\label{eq5}
    Y_{mask}^{(t)} = \mathop{\arg\min}\limits_{y_{i}\in Y_{mask}^{(t-1)}}(p_{i},N_{t})
\end{equation}
\end{small}

\begin{small}
\begin{equation}\label{eq6}
    Y_{obs}^{(t)} = Y \backslash Y_{mask}^{(t)}
\end{equation}
\end{small}
Where $p_{i}$ is the probability score of $y_{i}$ calculated in Equation \ref{yi}, \ref{pi}.
$Y_{mask}^{(t)}$ 
is the set of masked tokens that are probably wrong at the $t_{th}$ iteration, and $Y_{obs}^{(t)}$ 
is the set of unmasked tokens that are considered correct and fixed in later iterations.  
At each iteration, the model predicts the probably wrong tokens in $Y_{mask}^{(t)}$ conditioned on the source
text $X$ and $Y_{obs}^{(t)}$. We select the prediction with the highest probability for each masked token $y_{i} \in Y_{mask}^{(t)}$, and update its probability score accordingly:

\begin{small}
\begin{equation}\label{yi}
    y_{i}^{(t)} = \mathop{\arg\max}\limits_{w \in V}P(y_{i}=w|X,Y_{obs}^{(t)})
\end{equation}
\begin{equation}\label{pi}
    p_{i}^{(t)} = \mathop{\max}\limits_{w \in V}P(y_{i}=w|X,Y_{obs}^{(t)})
\end{equation}
\end{small}
where $P(y_{i}=w|X,Y_{obs}^{(t)})$ is the conditional probability of $y_{i}$ being predicted as the token $w$ in the vocabulary set $V$.

\section{Experimental Setup}
\subsection{Dataset and Metrics}
\noindent\textbf{Training dataset}\quad
Following \citet{liu-etal-2021-plome}, the training data is composed of 10K manually annotated samples from SIGHAN \cite{conf/acl-sighan/WuLL13} and 271K
automatically generated samples from \cite{wang2018hybrid}.

\noindent\textbf{Evaluation dataset}\quad
Following previous works, the SIGHAN15 test dataset \cite{conf/acl-sighan/TsengLCC15} is used 
to evaluate the proposed model. Statistics of the used datasets please refer to Appendix \ref{data}.

\noindent\textbf{Evaluation Metrics}\quad 
We evaluate model performance of detection and correction at sentence-level, with accuracy, precision, recall and F1 scores. We evaluate these metrics using the script from  \citet{journals/corr/abs-2004-14166}~\footnote{https://github.com/ACL2020SpellGCN/SpellGCN}. Moreover, following \citet{liu-etal-2021-plome},  we also report the sentence-level results evaluated by
SIGHAN official tool~\footnote{http://nlp.ee.ncu.edu.tw/resource/csc.
html}.

\begin{table*}[t]
\centering
\setlength{\tabcolsep}{4mm}
\small
\begin{tabular}{l|cccc|cccc}
    \cmidrule[\heavyrulewidth]{1-9}
    \multirow{2}{*}{Method} & & \multicolumn{2}{c}{Detection Level} & & & \multicolumn{2}{c}{Correction Level} \\
    \cline{2-9} & Acc. & Pre. & Rec. & F1. & Acc. & Pre. & Rec. & F1 \\
    \hline
    Confusionset \citeyearpar{conf/acl/WangTZ19} & - & 66.8 & 73.1 & 69.8 & - & 71.5 & 59.5 & 64.9 \\
    FASPell \citeyearpar{2019FASPell} & 74.2 & 67.6 & 60.6 & 63.5 & 73.7 & 66.6 & 59.1 & 62.6 \\
    SpellGCN \citeyearpar{journals/corr/abs-2004-14166} & - & 74.8 & 80.7 & 77.7 & - & 72.1 & 77.7 & 75.9 \\
    Chunk2020 \citeyearpar{2020Chunk} & 76.8 & 88.1 & 62.0 & 72.8 & 74.6 & 87.3 & 57.6 & 69.4  \\
    SM BERT \citeyearpar{conf/acl/ZhangHLL20} & 80.9 & 73.7 & 73.2 & 73.5 & 77.4 & 66.7 & 66.2 & 66.4 \\
    RoBERTa-DCN \citeyearpar{wang2021dynamic} & - & 76.6 & 79.8 & 78.2 & - & 74.2 & 77.3 & 75.7 \\
    ECSpell \citeyearpar{2022General} & 83.4 & 76.4 & 79.9 & 78.1 & 82.4 & 74.4 & 77.9 & 76.1 \\
    \hline
    PLOME* \citeyearpar{liu-etal-2021-plome} & - & 77.4 & 81.5 & 79.4 & - & 75.3 & 79.3 & 77.2 \\
    REALISE* \citeyearpar{2021Read} & 84.7 & 77.3 & 81.3 & 79.3 & 84.0 & 75.9 & 79.9 & 77.8 \\
    PHMOSpell* \citeyearpar{2021PHMOSpell} & - & \textbf{90.1} & 72.7 & 80.5 & - & \textbf{89.6} & 69.2 & 78.1  \\
    MLM-phonetics* \citeyearpar{zhang2021correcting} & - & 77.5 & \textbf{83.1} & 80.2 & - & 74.9 & \textbf{80.2} & 77.5  \\
    GAD* \citeyearpar{guo2021global} & - & 75.6 & 80.4 & 77.9 & - & 73.2 & 77.8 & 75.4 \\
    SpellBert* \citeyearpar{ji2021spellbert} & - & 87.5 & 73.6 & 80.0 & - & 87.1 & 71.5 & 78.5 \\
    \hline
    BERT-finetune & 82.4 & 74.2 & 78.0 & 76.1 & 81.0 & 71.6 & 75.3 & 73.4  \\
    Our EGCM & 86.4 & 82.7 & 77.6 & 80.0 & 85.8 & 80.6 & 74.7 & 77.5 \\
    Pre-Tn EGCM* & \textbf{87.2} & 83.4 & 79.8 & \textbf{81.6} & \textbf{86.3} & 81.4 & 78.4 & \textbf{79.9} \\
    \cmidrule[\heavyrulewidth]{1-9}
\end{tabular}
\caption{\label{mainresults} Performance on the SIGHAN15 test set. Best results are in bold. The first group lists the models that are not pretrained, and the second group lists the methods that are pretrained (denoted with "*" ). } 
\end{table*}

\begin{table}
\setlength{\tabcolsep}{1.5mm}
\centering
\small
\begin{tabular}{c|ccc|ccc} 
    \toprule
    \multirow{2}{*}{Method} & \multicolumn{3}{c|}{Detection level} & \multicolumn{3}{c}{Correction level} \\
    & Pre & Rec & F1 & Pre & Rec & F1 \\ 
    \hline
    SpellGCN & 85.9 & \textbf{80.6} & 83.1 & 85.4 & 77.6 & 81.3 \\
    ECSpell & 85.7 & 78.4 & 81.9 & 85.4 & 76.6 & 80.7 \\
    TtT & 85.4 & 78.1 & 81.6 & 85.0 & 75.6 & 80.0 \\
    GAD & 86.0 & 80.4 & 83.1 & 85.6 & \textbf{77.8} & 81.5 \\
    Pre-Tn EGCM & \textbf{93.5} & 76.7 & \textbf{84.3} & \textbf{91.4} & 74.5 & \textbf{82.1} \\ 
    \bottomrule
\end{tabular}
\caption{\label{official} Performance on the SIGHAN15 test evaluated
by the official tools. Best results are in bold.}
\end{table}

\subsection{Comparing Methods}
We compare the performance of our model with several strong baseline methods as follows:

\noindent\textbf{Confusionset} introduces a copy mechanism into seq2seq and generates characters from the confusionset \cite{conf/acl/WangTZ19}.

\noindent\textbf{FASPell} utilizes a denoising autoencoder to generate candidates  \cite{2019FASPell}.

\noindent\textbf{SpellGCN} incorporates phonological and visual knowledge via a graph convolutional network \cite{journals/corr/abs-2004-14166}.

\noindent\textbf{Chunk} proposes a chunk-based decoding method
with global optimization \cite{2020Chunk}.

\noindent\textbf{SM BERT} uses soft-masking technique to connect the network of detection and correction \cite{conf/acl/ZhangHLL20}.

\noindent\textbf{TtT} employs a Transformer Encoder with a Conditional Random Fields layer stacked \cite{2021Tail}.

\noindent\textbf{PLOME} proposes a confusion set based masking strategy \cite{liu-etal-2021-plome}.

\noindent\textbf{REALISE} leverages the multimodal information and mixes them electively \cite{2021Read}.

\noindent\textbf{PHMOSpell} integrates pinyin and glyph with a multi-modal
method \cite{2021PHMOSpell}.

\noindent\textbf{ECSpell} adopts the Error Consistent masking strategy for pretraining \cite{2022General}.

\noindent\textbf{MLM-phonetics} integrates phonetic features by leveraging pre-training and fine-tuning \cite{zhang2021correcting}.

\noindent\textbf{RoBERTa-DCN} generates the candidates via a Pinyin Enhanced Generator \cite{wang2021dynamic}. 

\noindent\textbf{SpellBert} employs a graph neural network to introduce visual and phonetic features \cite{ji2021spellbert}.

\noindent\textbf{GAD} learns the global relationships of the potential correct input characters and the candidates of potential error characters \cite{guo2021global}.

\noindent\textbf{BERT} We also implement classical methods for comparison. We fine-tune the Chinese BERT model \cite{2018BERT} on the CGEC corpus directly.

\subsection{Hyperparameter Setting}
We follow most of the standard hyperparameters for transformers in the base configuration \cite{vaswani2017attention} and follow
the weight initialization scheme from BERT \cite{2018BERT}. For regularization, we use 0.3 dropout, 0.01 L2
weight decay. The hyperparameter $\gamma$ which is used to weight
the confusion loss is set to 2 after tuning. Adam optimizer \cite{2014Adam} with
$\beta = (0.9, 0.999), \varepsilon = 1e^{-6}$ is used to conduct the parameter learning. The learning rate is set to $5e^{-5}$, and the model is trained with learning rate
warming up and linear decay.

\section{Results and Analysis}
\begin{table}[t]
\setlength{\tabcolsep}{1mm}
\centering
\small
\begin{tabular}{l|ccc|ccc}
    \toprule
    \multirow{2}{*}{Method} & \multicolumn{3}{c|}{Detection Level} & \multicolumn{3}{c}{Correction Level}  \\
    & Pre. & Rec. & F1. & Pre. & Rec. & F1. \\
    \hline
    EGCM & 82.7 & 77.6 & 80.0 & 80.6 & 74.7 & 77.5 \\
    - EFEnc & 80.5 & 75.2 & 77.8 & 78.7 & 72.3 & 75.4 \\
    - CFL & 79.2 & 72.7 & 75.8 & 77.3 & 69.9 & 73.4 \\
    - GFI & 77.5 & 76.9 & 77.2 & 77.1 & 73.9 & 75.7 \\
    \bottomrule
\end{tabular}
\caption{\label{ablation} Ablation study on SIGHAN15. "-EFEnc" means removing the error-focused encoder. "-CFL" means removing the confusion set loss.
"-GFI" means not using \textit{Guidance for Inference} in the inference stage.} 
\end{table}

\begin{table}[t]
\small
\setlength{\tabcolsep}{1.5mm}
\centering
\begin{tabular}{c|c|c} 
    \toprule
    top-k   & $P_{error\&mask/error}$ & $P_{correct/unmask}$ \\ 
    \hline
    k=1 & 94\% & 99.8\% \\
    k=2 & 90\% & 99.7\% \\
    k=3 & 88\% & 99.7\% \\
    \bottomrule
\end{tabular}
\caption{\label{eva}An evaluation on Zero-shot error detection.}
\end{table}

\subsection{Overall Performance}
Table \ref{mainresults} reports the performance of our proposed EGCM model and baseline models on the SIGHAN15 test set. 
For a fair comparison, we also employ the pre-trained model cBERT \cite{liu-etal-2021-plome} which has the same architecture with BERT and pre-trained via the confusion set based masking strategy.
Our model with pretrained cBERT (Pre-Tn EGCM) outperforms all existing approaches, achieving a 81.6 F1 at detection and 79.9 F1 at correction. Compared with the BERT baseline, Pre-Tn EGCM achieves 5.5\% performance gain on detection F1 and 6.5\% gain on correction F1. Among un-pretrained methods, EGCM also outperforms all competitor models
by a wide margin. 

We also evaluate the model performance using the official tool, and report the results in Table \ref{official}. Our model Pre-Tn EGCM obtains the best results for both detection and correction. Especially, it greatly outperforms previous methods in precision.  

It should be emphasized that, our model EGCM is trained on 270k HybirdSet and outperforms several models that are pre-trained on a big size of synthetic 
data, such as PLOME  \cite{liu-etal-2021-plome} which is pre-trained using 162 million sentences. This demonstrates that our model effectively learns to correct spelling errors without relying on heavyweight data.
An example output of our EGCM comparing with BERT is listed in Appendix \ref{case study}.
\subsection{Ablation Study}
We explore the contribution of each component in our EGCM model by conducting ablation studies with the following settings: \textbf{(1)} Removing the error-focused encoder mentioned in \ref{ewe}. \textbf{(2)} Removing the confusion set loss $L_{cs}$ in equation \ref{eq2}. \textbf{(3)} Initialize the start sequence of inference with all \texttt{[MASK]} instead of using the \textit{Guidance for Inference}. The results are shown in Table \ref{ablation}. 

Specifically, the confusion set loss leads to the biggest improvement to our model with 
4.2 points for detection and 4.1 points for correction.  
By removing the error-focused encoder, the drop of performance indicates that this encoder does learn to pay attention to the probably wrong tokens of the sentence and impel our model to correct the wrong tokens actively. Also, without the use of \textit{Guidance for Inference} as the start of decoding for inference, the performance drops especially on precision, which indicates that by fixing the tokens that are correct can effectively avoid 
over-correction and improve precision. 

\subsection{Evaluation on Zero-shot Error Detection}
We employ a zero-shot detection approach to do a preliminary detection, in which all the tokens are divided into two groups, the probably wrong tokens and the probably correct one. In the inference stage, the probably correct ones are unmasked and will not be modified to avoid over-correction, while the probably wrong ones are masked and repredicted
. We want to ensure that unmasked tokens are truly correct that don't need to be modified, and at the same time, the errors in the sentences are masked as many as possible. 

As shown in Table \ref{eva}, $P_{error\&mask/error}$ denotes the percentage of errors that are masked, $P_{correct/unmask}$ denotes the percentage of truly correct tokens in the unmasked tokens. In our zero-shot error detection, the BERT predicted tokens with top-$k$ probabilities are selected as candidates, if the original token is not in the candidates list, it is considered as wrong. We try different $k$ and conduct experiments. Our method achieves promising results with high accuracy, which guarantees correct signals for further processing. Obviously, the smaller $k$ is, the more tokens are masked and less tokens are fixed, this might lead to over-correction. We want the errors are masked as many as possible, and at the same time, fewer tokens are masked. Therefore, in our model, we set $k=2$. 

\subsection{Analysis on Different Confusion Sets}
\begin{table}[t]
\setlength{\tabcolsep}{0.5mm}
\small
\centering

\newcommand{\tabincell}[2]{\begin{tabular}{@{}#1@{}}#2\end{tabular}}
\begin{tabular}{c|c|ccc|ccc} 
    \toprule
    \multirow{2}{*}{\tabincell{c}{Confusion\\set}} & \multirow{2}{*}{Method} & \multicolumn{3}{c|}{Detection level} & \multicolumn{3}{c}{Correction level}  \\
    &  & Pre. & Rec. & F1. & Pre. & Rec. & F1.                     \\ 
    \hline
    \multirow{2}{*}{\tabincell{c}{[1] \\ \citeyearpar{2022General}}}
    & ECSpell & 76.4 & 79.9 & 78.1 & 74.4 & 77.9 & 76.1 \\
    & EGCM & 82.7 & 77.6 & 80.0 & 80.6 & 74.7 & 77.5 \\ 
    \hline
    \multirow{4}{*}{\tabincell{c}{[2] \\ \citeyearpar{conf/acl-sighan/WuLL13}}}
    & SpellGCN & 74.8 & 80.7 & 77.7 & 72.1 & 77.7 & 75.9 \\
    & EGCM & 80.6 & 78.2 & 79.4 & 78.2 & 76.3 & 77.2 \\ 
    \cline{2-8}
    & PLOME* & 77.4 & 81.5 & 79.4 & 75.3 & 79.3 & 77.2 \\
    & Pre-Tn EGCM* & 81.6 & 79.6 & 80.6 & 79.8 & 76.4 & 78.1 \\ 
    \hline
    \multirow{2}{*}{\tabincell{c}{[3] \\ \citeyearpar{wang2018hybrid}}}
    & Confusionset & 66.8 & 73.1 & 69.8 & 71.5 & 59.5 & 64.9 \\
    & EGCM & 79.5 & 74.7 & 77.0 & 77.4 & 71.7 & 74.4 \\
    \bottomrule
\end{tabular}
\caption{\label{cs} Effects of different confusion sets. }
\end{table}
To further prove the effectiveness of the confusion loss we proposed, and to show that this loss function can be generalized, we conduct experiments on three different confusion sets, 
%
including the confusion set proposed by \citet{2022General}~\footnote{https://github.com/Aopolin-Lv/ECSpell},  \citet{conf/acl-sighan/WuLL13}~\footnote{http://nlp.ee.ncu.edu.tw/resource/csc.html}, and \citet{wang2018hybrid}~\footnote{https://github.com/wdimmy/Automatic-Corpus-Generation}. 
For each confusion set, we compare our model with the models that use the same confusion set but in different way.  

As shown in table \ref{cs}. 
For all three confusion sets, our model outperforms the model that utilizes the same confusion set. Compared with previous methods, our model takes every token in the confusion set into consideration by computing it's possibility of being misused. Moreover, the results indicate that our model has strong generalization ability and is not limited to any specific confusion set.



\subsection{Analysis on Decoding Iterations}

\begin{figure}[h] 
\centering
\includegraphics[width=0.3\textwidth]{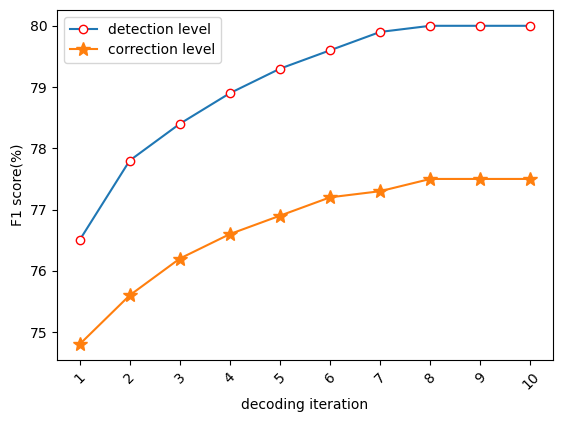} 
\caption{Results of different decoding iterations} 
\label{dc} 
\end{figure}

With a predefined decoding iteration $T=10$, we show the F1 score of previous iterations $t(t<T)$ to illustrate how the mask-predict strategy detects and corrects the wrong tokens step by step. As shown in figure \ref{dc}, 
the F1 score of detection and correction improves as the decoding iteration goes up. This indicates that, by masking and repredicting the tokens of low probability in each iteration, our model corrects the tokens that are wrongly predicted during previous iterations. And as the number of unmasked tokens increases, more information is given to help predict the hard masked tokens. With 8 iterations, our model achieves state-of-the-art performance. 

\begin{table}[t]
\setlength{\tabcolsep}{1.5mm}
\centering
\small
\begin{tabular}{l|c|c}
    \toprule
    Model & Time(ms) & Speedup  \\
    \hline
    Transfomer \citeyearpar{vaswani2017attention} & 63ms     & 1x       \\
    PLOME \citeyearpar{liu-etal-2021-plome}       & 45ms     & 1.4x     \\
    REALISE \citeyearpar{2021Read}                & 17ms     & 3.7x     \\
    TtT \citeyearpar{2021Tail}                    & 15ms     & 4.2x     \\
    EGCM                           & 10ms     & 6.3x     \\
    \bottomrule
\end{tabular}
\caption{\label{speed} Comparisons of the computing efficiency.}
\end{table}

\subsection{Analysis on Computing Efficiency}
Chinese spelling correction can be applied in many real-life applications, such as writing assistant and search engine.
Therefore, the time cost efficiency of models is a key point to be considered. 
We implement both the baseline models and our model on the single NVIDIA RTX 2080 GPU. Table \ref{speed} depicts the time cost per sample of our model comparing with some previous approaches. Our model runs faster than all previous approaches. 


\section{Conclusion}
We propose an error-guided correction model for the CSC task. A zero-shot error detection method is proposed to provide guidance signals for training and inference. We apply a conditional masked language model as a backbone, where we improve the encoder-decoder architecture by adding an error-focused encoder which pushes our model to focus on the wrong tokens. During training, we introduce a new confusion loss to help the model distinguish similar tokens. During inference, the error-guided mask-predict decoding strategy is adopted to mask and repredict the tokens that are probably wrong. Experimental results show that our model not only achieves superior performance against state-of-the-art approaches but also is cost-saving and green. 

\section*{Limitation}
In this paper, we use the results from zero-shot spelling error detection as a guidance signal. The sentence with probably wrong tokens masked and the other tokens fixed are used as a start of decoding. This means that if a wrong token is not assigned with a \texttt{[MASK]} token, it will never be corrected in later iterations. Even though we conduct experiments and the result shows that up to 94\% of the wrong tokens are masked in the guidance signal, there are still some wrong tokens missed by our model. To limit the number of tokens that are free to be modified is one of our ways to improve precision, but we are also looking forward to a way to further improve recall.  

What's more, even though we make full use of the confusion set, we still think that's not enough. Now we are using the confusion set in which every token has a set of predefined similar tokens. And these sets of similar tokens are isolated with each other. However, Chinese has various kinds of spelling errors, the target token might not be in the predetermined similar tokens set of the original token. And this kind of mistakes can never be learned to correct by the model. We think a better design for the data structure of the confusion set needs to be proposed, in which the sets are not isolated and we are able to calculate the similarity distance between each pair of tokens using particular algorithms, for example, UnionFind on a dynamic Graph. This kind of dynamic confusion knowledge can help avoid ignoring the probably misused tokens.

\section*{Acknowledgement}
This work is supported by the National Hi-Tech RD Program of China (No.2020AAA0106600),  
the National Natural  Science Foundation of  China (62076008) and the  Key Project of Natural Science Foundation of China (61936012).

\bibliography{anthology,custom}
\bibliographystyle{acl_natbib}

\appendix

\section{Statistics of Datasets}
\label{data}
\begin{table}[h]
\centering
\small
\begin{tabular}{llcl} 
\toprule
\textbf{Training Set} & \#Sent & Avg.Length & \#Errors  \\ 
\hline
SIGHAN13     & 700    & 41.8       & 343       \\
SIGHAN14     & 3437   & 49.6       & 5122      \\
SIGHAN15     & 2338   & 31.3       & 3037      \\
Wang271K     & 271329 & 42.6       & 381962    \\ 
\hline
\textbf{Test Set}     & \#Sent & Avg.Length & \#Errors  \\ 
\hline
SIGHAN15     & 1100   & 30.6       & 704       \\
\bottomrule
\end{tabular}
\caption{Statistics of used datasets.}
\end{table}

\section{Case Study}
\label{case study}
\begin{figure}[h]
\centering
    \includegraphics[width=0.4\textwidth]{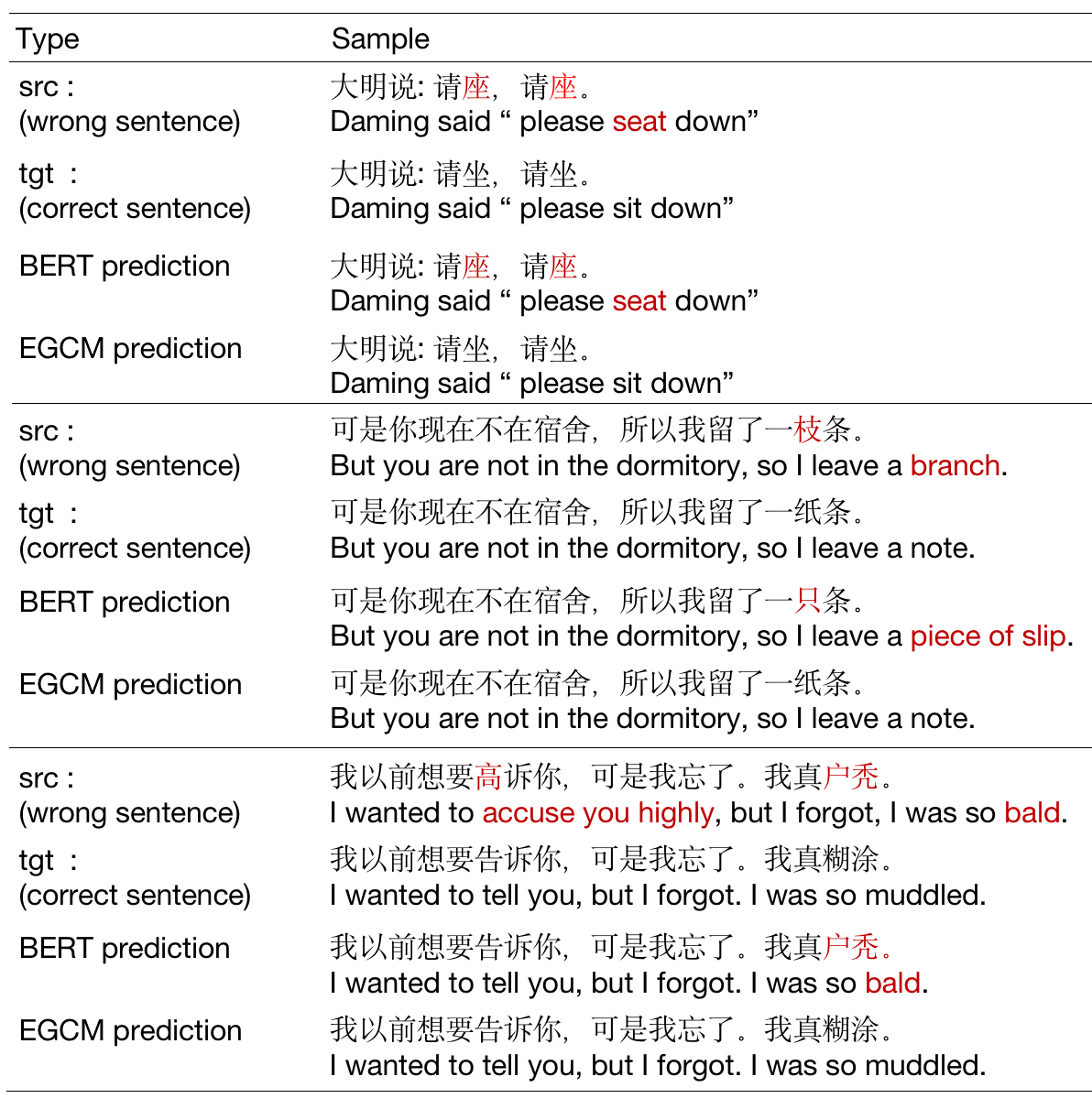} 
    \caption{Case Study. The wrong tokens are marked red. } 
    \label{case} 
\end{figure}

We list several cases of Chinese Spelling Correction. We present the source wrong sentence and the target correct sentence. We also present the corrections made by BERT and our EGCM.

\end{CJK}
\end{document}